\documentclass[ngerman,english]{bvm}

\bvmdef\type{V}
\bvmdef\articlenumber{3093} 

\date{}
\title{Heatmap-based 2D Landmark Detection with a Varying Number of Landmarks}

\titlerunning{Detection of a varying number of landmarks}

\author{Antonia~Stern$^{1}$, Lalith~Sharan$^1$, Gabriele~Romano$^2$, Sven~Koehler$^1$, Matthias~Karck$^2$, Raffaele~De~Simone$^2$, Ivo~Wolf$^3$, Sandy~Engelhardt$^1$}
\authorrunning{Stern et al.}
\institute{%
$^{1}$Group Artificial Intelligence in Cardiovasular Medicine (AICM), Department of Internal Medicine III, Heidelberg University Hospital, Heidelberg\\
$^{2}$Department of Cardiac Surgery, Heidelberg University Hospital, Heidelberg\\
$^{3}$Faculty of Computer Science, University of Applied Sciences, Mannheim}

\email{sandy.engelhardt@med.uni-heidelberg.de}
\begin{document}

%==============================================================================
% wählen Sie mit dem Befehl \selectlanguage die Sprache aus, in der Ihr 
% Proceeding verfasst ist
%
%\selectlanguage{german}
\selectlanguage{english}

\maketitle

\begin{abstract}
Mitral valve repair is a surgery to restore the function of the mitral valve. To achieve this, a prosthetic ring is sewed onto the mitral annulus. Analyzing the sutures, which are punctured through the annulus for ring implantation, can be useful in surgical skill assessment, 
for quantitative surgery and for positioning a virtual prosthetic ring model in the scene via augmented reality.
This work presents a neural network approach which detects the sutures in endoscopic images of mitral valve repair and therefore solves a landmark detection problem with varying amount of landmarks, as opposed to most other existing deep learning-based landmark detection approaches. 
The neural network is trained separately on two data collections from different domains with the same architecture and hyperparameter settings.
The datasets consist of more than $1,300$ stereo frame pairs each, with a total over $60,000$ annotated landmarks.
The proposed heatmap-based neural network achieves a mean positive predictive value (PPV) of $66.68\pm4.67\%$ and a mean true positive rate (TPR) of $24.45\pm5.06\%$ on the intraoperative test dataset and a mean PPV of $81.50\pm5.77\%$ and a mean TPR of $61.60\pm6.11\%$ on a dataset recorded during surgical simulation.
The best detection results are achieved when the camera is positioned above the mitral valve with good illumination. A detection from a sideward view is also possible if the mitral valve is well perceptible.
\end{abstract}

\section{Introduction}
Mitral valve reconstruction is a complicated surgery, where the surgeon implants a prosthetic ring to the tissue by placing approx. $12$ to $15$ sutures on the mitral annulus \cite{3093-01}.
Analysing how sutures are placed (e.g., their pattern and distances) can help in improving the quality and consistency of the procedure. This could be useful in surgery itself and in preoperative simulation for trainees \cite{3093-02}.
Besides that, the position of the sutures may be utilized to reconstruct the 3D position of the mitral annulus in a stereo-setting and a virtual ring model can be superimposed onto the endoscopic video stream prior to ring implantation \cite{3093-03,3093-04}.

Several deep learning-based methods exist for landmark detection in computer-vision and medical applications.
In general, they can be classified into heatmap approaches \cite{3093-05,3093-06}, coordinate regression  and patch-based \cite{3093-07} approaches.
Gilbert et al. \cite{3093-05} used a U-Net-like architecture to detect anatomical landmarks by predicting one heatmap per landmark.
The input label heatmaps contain 2D Gaussians centred at the annotated point location.
In contrast to that, coordinate regression approaches directly output the coordinates instead of heatmaps, which is less computationally expensive as no upsampling is necessary. 
Usually, these approaches consist of a sequence of convolutional layers along with a fully connected layer or a $1\times1$-convolution.
Another common approach is to detect landmarks in global and local patches with the help of displacement vectors \cite{3093-07}.

State-of-the-art literature on deep learning-based landmark detection focuses mainly on detecting a predefined number of keypoints. In medical applications, landmarks are often used to define a fixed number of anatomical points.
These approaches output one heatmap for each landmark or use a vector representation of fixed length with the length corresponding to the number of keypoints.
However, these methods cannot be applied to a setting with an arbitrary number of keypoints. Furthermore, in our scenario, unlike in most other works, the landmarks can lie in close proximity to each other, which makes the application of patch-based approaches difficult.

The work presented in this paper uses a network architecture that is able to deal with a varying number of landmarks, since, in principle, a varying number of sutures can be placed. This work uses a U-Net based architecture with a \textit{single} foreground heatmap as output. Thresholding this heatmap allows us to detect a non-predefined number of points in endoscopic images, which represent the positions where the sutures enter the tissue.

\section{Materials and methods}

\subsection{Network Architecture}
This work uses a U-Net-based \cite{3093-08} architecture with a depth of $4$. After each $3\times3$-convolution, batch normalisation is applied. The padded $3\times3$-convolutional layers use the \textit{ELU} activation function. The final layer is a $1\times1$-convolutional layer with a \textit{sigmoid} activation function. 
The first convolutional layer has $16$ filter maps, while the bottleneck layer has $256$ filter maps. The number of filters is doubled/halved after every maxpooling/upsampling in the downsampling/upsampling path. The dropout rate is set to values between $0.3$ and $0.5$ with the lowest dropout rate in the first downsampling/last upsampling block.
Unlike the original U-Net \cite{3093-08}, our architecture applies zero-padding in the convolutional layers. Therefore, the input size has to be divisible by $4^{depth}$, which equals $16$ in this work. The loss function is of the form $MSE - SDC$, where $MSE$ is the mean squared error function and $SDC$, the S{\o}rensen dice coefficient.

To preserve the aspect ratio of the images, a width of $512$ pixels and a height of $288$ pixels was chosen as the input size. The input images are RGB-images with $3$ channels. The two-channel output masks contain one channel for the suture landmarks and one for the background. This design is more efficient than heatmaps approaches, which use one channel for each landmark. The predicted output heatmap is converted into a binary mask by thresholding. Then, the centre of mass for each region is determined as point of interest.

\subsection{Dataset}
Two instances of the same neural network approach proposed in this work were trained separately on two different data sub-collections (Tab. \ref{3093-datasets}, A.1 and B) consisting of endoscopic images of mitral valve repair. One data collection contains very heterogeneous \textit{intraoperative} images extracted from videos which were recorded during surgery (different view angles, light intensities, number of sutures, surgical instruments, prosthesis etc.). Different amounts of frames were annotated from these datasets, therefore an additional hold-out test set was created from surgeries with significantly less frames (A.2). The other \textit{simulation} data collection was extracted from videos of operations performed on an artificial mitral valve replica made of silicone, which was first introduced in \cite{3093-02} and used in surgical training  applications.

The endoscopic videos were recorded from an Image S1 stereo camera (Karl Storz SE \& Co. KG, Tuttlingen, Germany) in full HD resolution or larger at $25\ts {\rm fps}$. Frames were saved in top-down format (left image top, right image bottom). Relevant scenes were identified before extracting the frames from the videos and afterwards every $120th$ frame was extracted. In scenes with rapid changes, every $10th$ frame was extracted and in scenes with only few changes, every $240th$ frame was extracted. These characteristics were identified manually.

This work uses the two subimages of the stereo recording as separate input images to the network, thus the final number of frames for training and testing is twice as large as given in Tab. \ref{3093-datasets}. 
The intraoperative dataset used for training consists of $2654$ frames, which were extracted from $5$ surgeries  (mean 530,8 $\pm$ 213,6 frames per surgery).
An additional balanced intraoperative test dataset was  annotated, containing $200$ frames extracted from $4$ surgeries (mean 100 $\pm$ 0 frames per surgery). 
The simulator datasets consists of $2708$ frames extracted from $10$ simulated surgeries (mean 270,7 $\pm$ 77,7 frames per surgery).

The frames were manually annotated with the tool \textit{labelme} \cite{3093-09}. During annotation, corresponding suture points in the left and right image of the stereo pair are joined with a line. If a suture point is only visible in one of the subimages it is marked as a point. 
The ground truth heatmaps were then created by placing 2D Gaussians at the position of the landmarks. For a down-scaled input size of $512\times288$, a variance $\sigma$ of $1$ was used.

\begin{table}[t]
\centering
\caption{Three sub-collections from two different domains were used. Note that we treat left and right stereo frame independently in our work, therefore training frames are  doubled. The additional intraop test set has significantly lower number of labeled frames, hence we decided to not include it directly in the cross validation (CV).}
\label{3093-datasets}
\begin{tabular*}{\textwidth}{l|@{\extracolsep\fill}lccc}
\hline
domain             & usage & \# stereo frame & \# suture endpoints & \# surgeries \\ \hline
A.1 intraop  & 5-fold CV (train, test)  & $1,327$   & $26,937$  & $5$             \\
A.2 intraop  & test                     & $200$     & $4,305$   & $4$             \\
B simulator  & 5-fold CV (train, test)  & $1,354$   & $33,893$  & $10$            \\ \hline
\end{tabular*}
\end{table}

During training the images are randomly augmented with a probability of 80\% using tensorflow functions: 
rotation of $\pm60^\circ$, pixel shifting in a range of $\pm10\%$, mask pixel shifting in a range of $\pm1\%$,  shearing in a range of $\pm$0.1, brightness in a range of $\pm0.2$, contrast in a range from $0.3$ to $0.5$, random saturation in a range from $0.5$ to $2.0$ and hue in a range of $\pm0.1$.
Additionally the images were flipped horizontally and vertically with a probability of $50\%$.

\subsection{Evaluation}

During method development, which involved hyperparameter tuning, training and validation was performed on dataset A.1. The best neural network had a total of $2.1M$ trainable parameters. The initial learning rate was set to $10\textsuperscript{-3}$ with a decay factor of $0.1$. 
After fixing these parameters, final evaluation was conducted on A.1, A.2 and B, involving two 5-fold cross validations (CV) on different splits of dataset A.1 and B. Furthermore, an additional disjoint test set was used (A.2) to further assess method generalizability. To prevent data leakage, dataset splitting was always carried out on the level of the surgeries. The results are given for the epoch $\leq 200$ with the lowest value of the validation loss.
In total, $10$ different models were trained, one for each fold of each application (intraop and simulation).

A suture point detection is considered successful if the centres of mass of ground truth and prediction are less than $6$ pixels apart. On an image of size $512\times288$, this radius roughly corresponds to the thickness of a suture when it enters the tissue. 
If a point in the produced mask or the ground truth mask is assigned multiple times, only the matched pair with the least distance is kept. 
Every matched point from the produced mask is considered a true positive (TP). Predicted points that could not be matched to any ground truth point are defined as false positives (FP) and all ground truth points without a corresponding point in the produced mask are false negatives (FN).
To evaluate the precision (positive predictive value, PPV) and sensitivity (true positive rate, TPR) are used. Precision and sensitivity are defined as $PPV = TP/(TP+FP)$ and $TPR = TP/(TP+FN)$.
The values of PPV and TPR are displayed over a threshold from $0.05$ to $1.0$ to show the influence of thresholding the heatmap on the detection rate.

\section{Results}

\begin{figure}[t]
    \centering
    \subfigure[simulator dataset (B)]{
        \centering
        \includegraphics[width=0.18\textwidth]{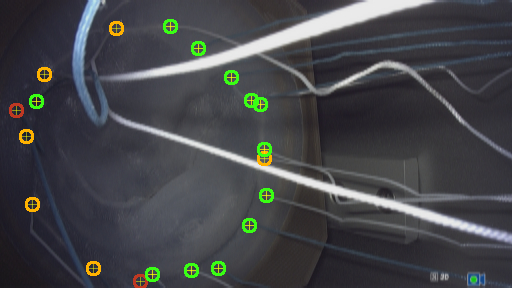} 
        \includegraphics[width=0.18\textwidth]{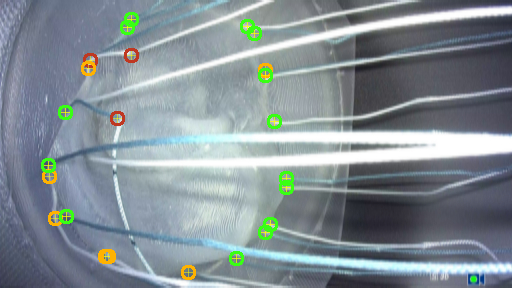} 
        \includegraphics[width=0.18\textwidth]{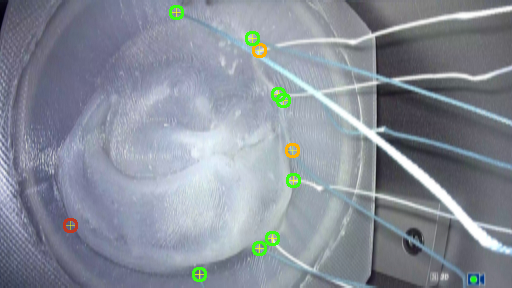}
        \includegraphics[width=0.18\textwidth]{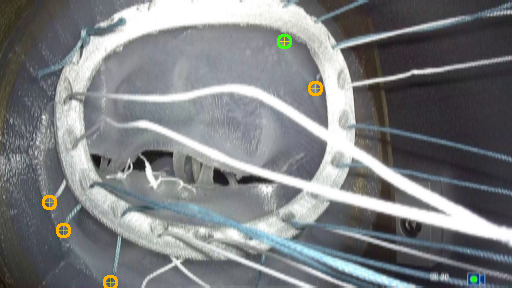} 
        \includegraphics[width=0.18\textwidth]{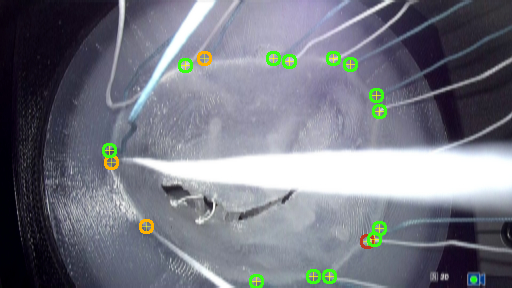} 
        \label{3093-predictions-simdata}}
        
    \subfigure[intraoperative test dataset (A.2)]{
        \centering
        \includegraphics[width=0.18\textwidth]{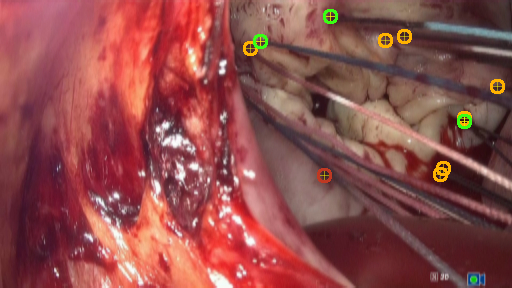}
        \includegraphics[width=0.18\textwidth]{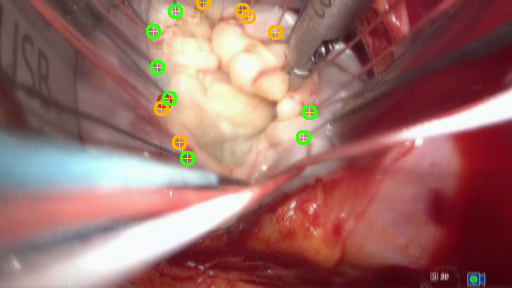}
        \includegraphics[width=0.18\textwidth]{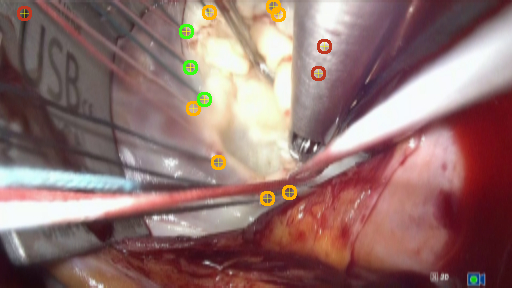} 
        \includegraphics[width=0.18\textwidth]{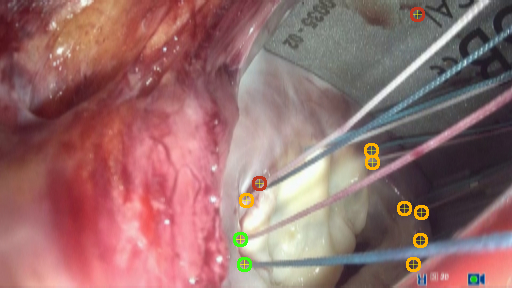}
        \includegraphics[width=0.18\textwidth]{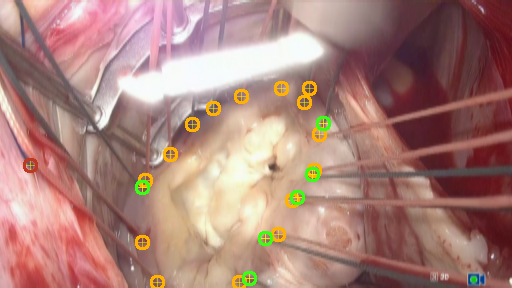} 
        \label{3093-predictions-intraop}}
	\label{3093-predictions}
	\caption{Example predictions on the two domains. Green circles represent true positives (TP), red circles show false positives (FP) and orange circles show false negatives (FN).}
\end{figure}

Some visual examples are provided in Fig. \ref{3093-predictions-simdata} and Fig.\ref{3093-predictions-intraop}. In intraoperative images, green sutures are better recognized than white ones because they can be better distinguished from the background. During surgery (not during simulation), white sutures often appear red because they are soaked with blood which makes it even harder to distinguish them from the background. 

\begin{figure}[b]
    \centering
    \subfigure[CV on A.1 (intraop)]{
        \centering
        \includegraphics[width=0.3\textwidth]{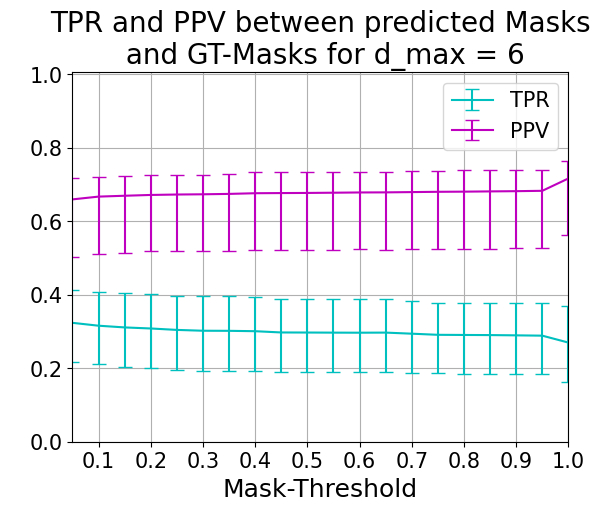} 
        \label{3093-result-train}}
    \subfigure[Test on A.2 (intraop)]{
        \centering
        \includegraphics[width=0.3\textwidth]{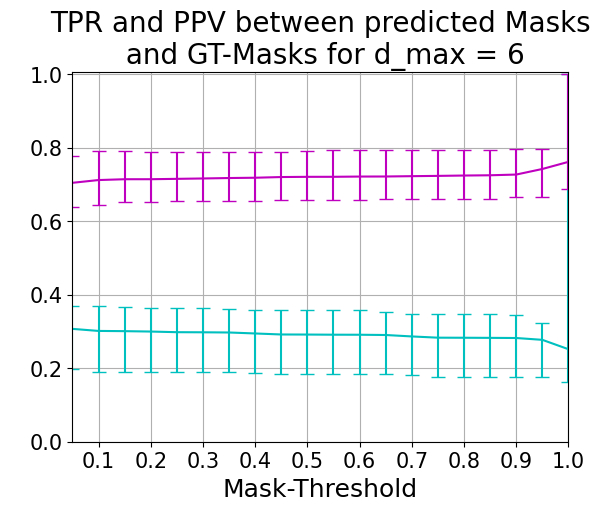}
        \label{3093-result-testset}}
    \subfigure[CV on B (sim)]{
        \centering
        \includegraphics[width=0.3\textwidth]{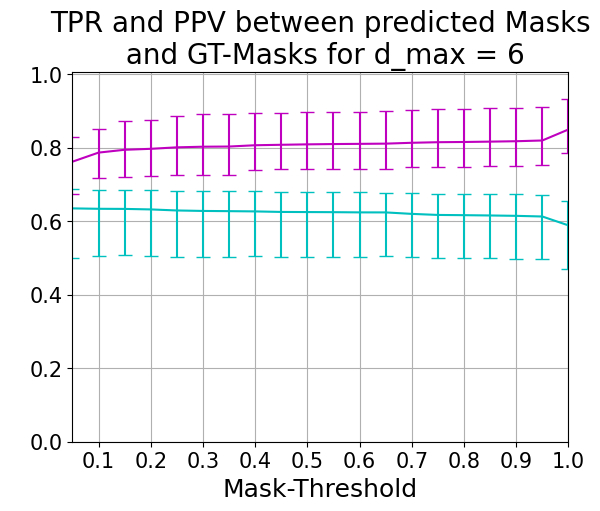}
        \label{3093-result-simdata}}
	\caption{The mean PPV and TPR of all folds with variable mask threshold. The bars show the value of the fold with the lowest and highest value respectively.}
	\label{3093-results}
\end{figure}

The model achieved a mean PPV of $67.99\pm7.69\%$ and a mean TPR of $29.03\pm7.74\%$ on the intraoperative dataset (A.1) for a threshold of $t=0.8$ during testing on the respective hold-out folds (Fig. \ref{3093-result-train}). 
All models from the CV were also applied to the separate test set (A.2) and comparable results were obtained with a mean PPV of $66.68\pm4.67\%$ and a mean TPR of $24.45\pm5.06\%$ (Fig. \ref{3093-result-testset}), meaning that it generalizes beyond the surgeries where hyperparameter tuning was performed on.
When training the model on the simulator dataset (B) with the same settings, a mean PPV of $81.50\pm5.77\%$ and a mean TPR of $61.60\pm6.11\%$ is achieved during CV (Fig. \ref{3093-result-simdata}). As can be seen in the plots, the estimated TPR and PPV are not sensitive with regard to the chosen threshold. 
Performance differences between the folds are expressed by the bars in Fig. \ref{3093-results}.

\section{Discussion} 

The developed approach is able to detect a varying number of landmarks, which represent the position of sutures stitched through the mitral annulus. Evaluation on a large dataset of two domains revealed that the neural networks detects sutures in most of the scenes, but had difficulties in cases where the mitral annulus is partly occluded by the prosthetic ring, the surgical tools or tissue (Fig. \ref{3093-predictions-simdata} $4th$ image).
Reflections or embossings on tools has also led to an increase in FPs (Fig. \ref{3093-predictions-intraop} $3rd$ image). 
The best detection results were achieved when the camera was positioned above the mitral valve with good illumination. 
Comparing both domains, the network performance was better on the simulation domain, where the camera angle is often more favorable and the silicone appearance is more homogeneous in comparison to intraoperative scenes with various tissue textures and blood. However, the performance differences could be also explained by the higher numbers of simulated surgery instances used during training in the simulator domain.

In this work, two neural networks were trained separately on two datasets from different domains with the same architecture and hyperparameter settings. 
Future work includes incorporating image-to-image translation by generative adversarial networks \cite{3093-10} to adapt between the domains and to allow for joint landmark detection in both domains. 

\ack{The research was supported by the German Research Foundation DFG Project 398787259, DE 2131/2-1 and EN 1197/2-1 and by Informatics for Life funded by the Klaus Tschira Foundation.}

\bibliographystyle{bvm}

\bibliography{3093}

\begin{thebibliography}{10}

\bibitem{3093-01}
Carpentier A, Adams D, Filsoufi F.
\newblock {C}arpentier's {R}econstructive {V}alve {S}urgery.
\newblock Saunders; 2010.

\bibitem{3093-02}
Engelhardt S, Sauerzapf S, Preim B, et~al.
\newblock Flexible and comprehensive patient-specific mitral valve silicone
  models with chordae tendinae made from {3D}-printable molds.
\newblock Int J Comput Assist Radiol Surg. 2019;14(7):1177--1186.

\bibitem{3093-03}
Engelhardt S, De~Simone R, Zimmermann N, et~al.
\newblock Augmented reality-enhanced endoscopic images for annuloplasty ring
  sizing.
\newblock In: Augmented Environments for Computer-Assisted Interventions.
  Springer International Publishing; 2014.  p. 128--137.

\bibitem{3093-04}
Engelhardt S, Kolb S, De~Simone R, et~al.
\newblock Endoscopic feature tracking for augmented-reality assisted prosthesis
  selection in mitral valve repair.
\newblock In: Proc {SPIE}, Medical Imaging: Image-Guided Procedures, Robotic
  Interventions, and Modeling. vol. 9786; 2016.  p. 402--408.

\bibitem{3093-05}
Gilbert A, Holden M, Eikvil L, et~al.
\newblock Automated left ventricle dimension measurement in 2D cardiac
  ultrasound via an anatomically meaningful CNN approach.
\newblock In: Smart Ultrasound Imaging and Perinatal, Preterm and Paediatric
  Image Analysis. Springer International Publishing; 2019.  p. 29--37.

\bibitem{3093-06}
Jin H, Liao S, Shao L.
\newblock Pixel-in-pixel net: towards efficient facial landmark detection in
  the wild.
\newblock arXiv:200303771v1 [csCV]. 2020;.

\bibitem{3093-07}
Noothout JMH, De~Vos BD, Wolterink JM, et~al.
\newblock Deep learning-based regression and classification for automatic
  landmark localization in medical images.
\newblock IEEE Trans on Med Imag. 2020; p. 1--1.

\bibitem{3093-08}
Ronneberger O, Fischer P, Brox T.
\newblock U-Net: convolutional networks for biomedical image segmentation.
\newblock In: MICCAI. vol. 9351 of LNCS. Springer; 2015.  p. 234--241.

\bibitem{3093-09}
Wada K. labelme: image polygonal annotation with python; 2016.
\newblock \url{https://github.com/wkentaro/labelme}.

\bibitem{3093-10}
Engelhardt S, Simone RD, Full PM, et~al.
\newblock Improving surgical training phantoms by hyperrealism: deep unpaired
  image-to-image translation from real surgeries.
\newblock In: MICCAI. Springer International Publishing; 2018.  p. 747--755.

\end{thebibliography}
% Bitte setzen Sie hier Ihre Beitragsnummer ein und benennen Sie
% die BibTeX-Datei ebenfalls auf Ihre Beitragsnummer um.
%Kontrollzeiledef
\marginpar{\color{white}E\articlenumber} % Zeile nicht verändern!
\end{document}